\documentclass[letterpaper, 10 pt, conference]{ieeeconf}  % Comment this line out if you need a4paper

\IEEEoverridecommandlockouts                              % This command is only needed if 
\usepackage{graphics} % for pdf, bitmapped graphics files
\usepackage{epsfig} % for postscript
\usepackage{mathtools}
\usepackage{amsmath} % assumes amsmath package installed
\usepackage{amssymb}  % assumes amsmath package installe
\usepackage{algorithm}
\usepackage{algorithmic}
\usepackage{todonotes}
\usepackage{xcolor}
\usepackage{graphicx}
\usepackage{caption,subcaption}
\newcommand\colorcomment[1]{\textcolor{red}{#1}}
\usepackage{multirow}
\usepackage{mathrsfs}
\usepackage{ctable} % for \specialrule command
\usepackage{url}

\newcommand{\R}{\mathbb{R}}

\usepackage{array,booktabs,ragged2e}
\newcolumntype{R}[1]{>{\RaggedLeft\arraybackslash}p{#1}}

\DeclareMathOperator*{\argmax}{argmax}
\DeclareMathOperator*{\argmin}{arg\,min}
\usepackage{kotex}

\usepackage{tikz}

\newcommand\copyrighttext{%
  \footnotesize \textcopyright 2021 IEEE.  Personal use of this material is permitted.  Permission from IEEE must be obtained for all other uses, in any current or future media, including reprinting/republishing this material for advertising or promotional purposes, creating new collective works, for resale or redistribution to servers or lists, or reuse of any copyrighted component of this work in other works.}
\newcommand\copyrightnotice{%
\begin{tikzpicture}[remember picture,overlay]
\node[anchor=south,yshift=10pt] at (current page.south) {\fbox{\parbox{\dimexpr\textwidth-\fboxsep-\fboxrule\relax}{\copyrighttext}}};
\end{tikzpicture}%
}
\title{\LARGE \bf
Topology-Guided Path Planning for Reliable Visual Navigation of MAVs}

\author{Dabin Kim$^{*}$, Gyeong Chan Kim$^{*}$, Youngseok Jang, and H. Jin Kim% <-this % stops a space
\thanks{
This work was supported by Institute of Information \& Communications Technology Planning \& Evaluation(IITP) grant funded by the Korea government(MSIT) (No. 2019-0-00399, Development of A.I. based recognition, judgement and control solution for autonomous vehicle corresponding to atypical driving environment)
\newline 
\indent Dabin Kim, and Youngseok Jang are with the Mechanical and Aerospace Engineering Department, Seoul National University, Seoul 08826, South Korea (e-mail: dabin404@snu.ac.kr, duscjs59@gmail.com).
\newline
\indent Gyeong Chan Kim and H. Jin Kim are with the Aerospace Engineering Department, Seoul National University, Seoul 08826, South Korea (e-mail: skykim0609@snu.ac.kr, hjinkim@snu.ac.kr, corresponding author: H. Jin Kim).
\newline
$^{*}$ These authors contributed equally to this manuscript.}%
}

\begin{document}

\maketitle
\copyrightnotice
\thispagestyle{empty}
\pagestyle{empty}

\vspace{-3.5mm}
%%%%%%%%%%%%%%%%%%%%%%%%%%%%%%%%%%%%%%%%%%%%%%%%%%%%%%%%%%%%%%%%%%%%%%%%%%%%%%%%
\begin{abstract}
%\textcolor{red}{Aerial Robitcs session에 제출하고 싶어서 제목에 MAV를 추가하였습니다. }
Visual navigation has been widely used for state estimation of micro aerial vehicles (MAVs). For stable visual navigation, MAVs should generate perception-aware paths which guarantee enough visible landmarks. Many previous works on perception-aware path planning focused on sampling-based planners. However, they may suffer from sample inefficiency, which leads to computational burden for finding a global optimal path. To address this issue, we suggest a perception-aware path planner which utilizes topological information of environments. Since the topological class of a path and visible landmarks during traveling the path are closely related, the proposed algorithm checks distinctive topological classes to choose the class with abundant visual information. 
Topological graph is extracted from the generalized Voronoi diagram of the environment and initial paths with different topological classes are found. To evaluate the perception quality of the classes, we divide the initial path into discrete segments where the points in each segment share similar visual information. The optimal class with high perception quality is selected, and a graph-based planner is utilized to generate path within the class.
With simulations and real-world experiments, we confirmed that the proposed method could guarantee accurate visual navigation compared with the perception-agnostic method while showing improved computational efficiency than the sampling-based perception-aware planner.
\end{abstract}

%%%%%%%%%%%%%%%%%%%%%%%%%%%%%%%%%%%%%%%%%%%%%%%%%%%%%%%%%%%%%%%%%%%%%%%%%%%%%%%%
\section{INTRODUCTION} \label{label_intro}

Fully automated robotic operation requires a perception module that recognizes surrounding environment and estimates the robot state. In particular, visual odometry (VO) and simultaneous localization and mapping (SLAM) using vision sensors have been conducted for self-localization of micro aerial vehicles (MAVs) due to low weight, cost, and small size of the sensors while capturing abundant visual information of surrounding environments.
% \ys{ }{앞에 문장을 In particular, visual odometry (VO) and simultaneous localization and mapping (SLAM) using vision sensors have been conducted for self-localization for micro aerial vehicles (MAV) due to low weight, cost, and small size of the sensors while capturing abundant visual information of surrounding environments. 로 바꾸는건 어때?} 
% \ys{However}{however로 뒷문장이랑 연결하는거 좀 어색한듯 ㅠ, `이러한 vision-based perception modules의 성능은 dependent on the path-로봇이-움직이는 때문에, path를 계획할때 perception module의 고려는 필수적이다?.' 이런식으로 문장 구성해도 좋을거 같은데? },    
The integration of motion planning and perception modules poses significant issues for consideration, one of which is that the robot's localization capabilities are dependent on the path chosen by the robot. Therefore, it is necessary to take the perception module into account at the motion planning level, and this approach is called perception-aware motion planning. 
% \ys{ }{이러한 연구를 perception-aware path planning이라 부른다. 이거 여기 추가되도 좋은듯}

% \ys{Considering this situation, we studied a planning algorithm which considers the quality of the state estimation in the process of MAV moving from pre-defined start to goal point.}{요 문장 먼가 논문에 들어가기 좀 그래ㅠㅠ} 
In general, the performance of visual navigation is affected by the number and distribution of salient keypoints in the observed images. For example, if the generated trajectory passes through a texture-less area, it may become difficult to execute given missions due to accumulated error of state estimation.
Therefore, instead of conventional planners that are agnostic to the performance of the perception module, perception-aware motion planning is needed to lead stable visual navigation of MAVs.
%\ys{}{나는 robust나 stable 둘 중에 암거나 써도 좋을거 같애!} 

To this end, the previous perception-aware planners have  focused on approximation of navigation performance by inserting a perception-related cost on 
%{\color{blue}insert  as a perception-related cost to} (have 에 걸리는 거면 inserted 근데 그렇다면 insert 의 목적어가 또 필요 !! inserted a perception-related cost 의 의도아닌지?? )
existing motion planning algorithms. Especially, sampling-based methods \cite{achtelik2014motion}, \cite{sadat2014feature}, \cite{costante2018exploiting}
have been mainly used for perception-aware planners. However, since the evaluation of perception quality itself is usually computationally heavy \cite{makarenko2002experiment}, difficulty arises when the planner suffers from sample inefficiency in large environments.   
%(perception quality 좋은지 아닌지 계산하는거 자체가 복잡하다는 뜻 ??) {\color{blue}}

\begin{figure}[t]
    \centering
    \includegraphics[width=0.99\linewidth]{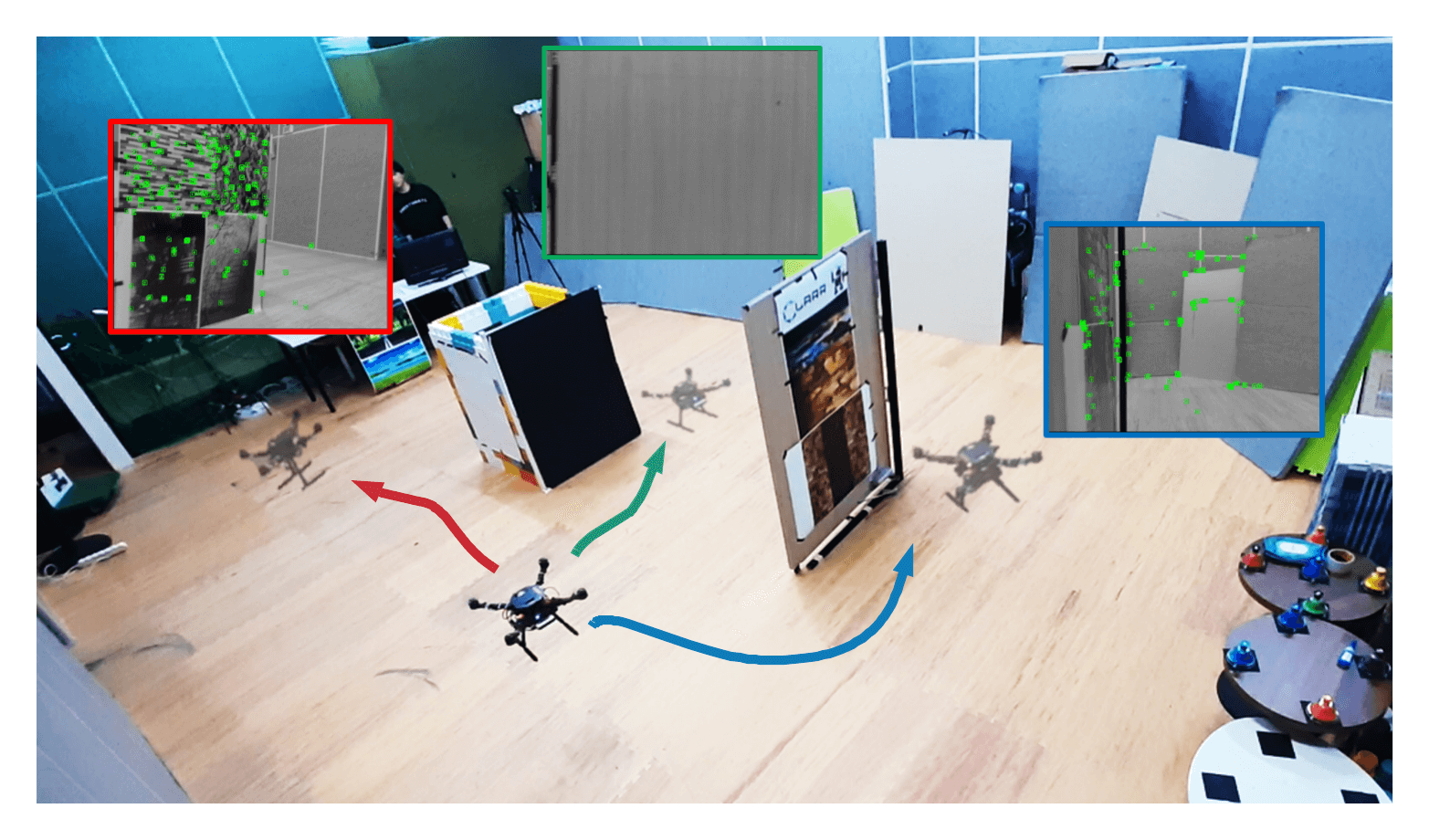}
    \caption{Snapshot from the experiment. For robust visual navigation, a micro aerial vehicle should decide a path which can guarantee enough visible features from multiple path candidates.}
    \label{fig:thumbnail}
\end{figure}

%Therefore, \ys{as an alternative of sampling-based planner, we studied topological planner which can exploit higher-level information about the environment.}{이러한 크고 복잡한 환경에서는 환경을 샘플 포인트에서 평가하기 보다는 higher-level information about the environment을 담고 있는 topology로 평가하는게 더 효율적이다??} 
Instead of covering the entire environment with randomly drawn samples, exploiting higher-level information about the environment improves efficiency of global planning. 
The topology of the environment is determined from the geometrical distribution of obstacles. This fact leads to an important heuristic in the perspective of visual navigation. The visibility of the landmarks is limited by the relative pose between MAV and obstacles, thus the topological class of path restricts the maximum obtainable visual information  \cite{fogliaroni2009qualitative}, \cite{stein2020enabling}. Therefore, taking topological information of paths into account helps to find whether the path is advantageous for visual navigation. In addition, once a reference path is generated, it cannot be updated to belong to a different topology through gradient-based optimization \cite{lavalle2006planning}. As a result, finding topological classes of paths with enough visual features gives important prior information for perception-aware path planning.
% \textcolor{blue}{it is necessary to find a good reference path which can acquire enough visual information for localization of the robot.} \textcolor{purple}{As a result, it is necessary to find a reference path within path topology which allows the MAV to acquire enough visual information for localization. - "path topology"를 대체하는 방향으로 문장 수정}
% \ys{}{나한테만 그런거일수 있는데, 이 단락내에서 흐름이 끊기는 부분이 있는거 같고 쉽게 안읽혀, 글고 경로 토폴로지에 대한 명확한 정의?가 단락 앞부분에서 필요해보여!}

In this paper, a topological perception-aware planner is suggested to prevent situations difficult to obtain accurate state estimation due to limited visual information. To create a path which attains both short path length and good perception quality, we propose the process of generating paths belonging to distinct topology classes and evaluate each path's quality with respect to path length and visual information.
% " Based on environment" 라는 phrase 이상한거 같음. 
% \colorcomment{Since whole validation and methods mainly focus on "totally" known environment, I think it is unnecessary to include "partially"} 
According to the authors' knowledge, this work is the first research on perception-aware motion planning that incorporates visual information with topological planning. The proposed planner can be exploited as an reference path to generate a feasible trajectory via trajectory optimization, or to provide prior information for a lower-level global planner. 
% Integrating with trajectory optimization method, we performed simulations in photorealistic environments and real-world experiments. We validated the proposed method to show higher success rates on state estimation than perception-agnostic planner, and confirmed that it is more computationally efficient than sampling-based planners. 
% \ys{}{이 문단에서, the proposed planner can ~ 이 부분만 살려서 앞 문단이랑 합쳐도 좋을거 같애. intro에서 뒤에 어떤 실험하고 validate 하다는거 안보여줘도 좋을거 같애서! abstract에서 적어주자 뒷부분은!}
% \textcolor{red}{이 부분에 Contribution을 강조하기 위해서 시뮬&실험 결과에 대해서 언급하는 것이 유익할까요?}

\section{RELATED WORK}
 \subsection{Perception-aware Planning}
Perception-aware motion planning refers to the algorithms that generate motion by considering the localization quality of the onboard navigation system. This study focuses specifically on motion planning algorithms that are applicable to visual navigation systems. In order to find a path from the start to the goal point, most perception-aware global planning algorithms are based on sampling-based methods. \cite{bry2011rapidly} suggested the Rapidly-exploring random belief tree (RRBT) for planning in belief space with a linear estimator based on a sampling-based method. As an extension of RRBT for vision systems, \cite{achtelik2014motion} applied local bundle adjustment (BA) in offline to estimate the covariance of future pose, and \cite{sadat2014feature} designed perception score based on the feature numbers in images as an approximation of localization quality. \cite{costante2018exploiting} utilized the photometric information of images for dense VO and biased the path toward texture-rich region. Unlike aforementioned methods, our work suggests a global planning method which can 
utilize the environment's topological information as a heuristic for perception quality.

Another focus of perception-aware motion planning is improving perception quality of the trajectory via considering tracking and triangulation of local landmarks. 
\cite{falanga2018pampc} formulated an optimal control problem which simultaneously minimizes the energy and velocity of the point of interest in image plane. \cite{murali2019perception} suggested differentiable cost for keeping visible features inside the Field of View (FoV) of the camera. \cite{bartolomei2020perception} applied feature triangulation and covisibility-related costs to gradient-based trajectory optimization, encouraging the tracking of local landmarks. \cite{zhang2018perception} and \cite{jang2020navigation} used receding-horizon method with designed planning cost regarding perception quality, which is evaluated from local landmarks.

\subsection{Topological Planning} 
Among other global planning methods such as sampling-based methods and search-based methods, topological planning methods obtain paths from the  topological graph of the environment. 
% Using sparse topological graph structure for path planning is memory-efficient \cite{collins2020efficient}, and simple low cost graph-search algorithms are sufficient to find connected path. Also, because path is divided into distinct topology classes, it can include higher level information such as visibility of obstacles \cite{jaillet2008path} and connectivity between regions \cite{blochliger2018topomap}. 
% Topological planner is also beneficial in finding global optimal path efficiently, similar to the purpose of this study.  
Topological planning is widely used in motion planning in the sense of reducing planning dimension with topological constraints \cite{stein2020enabling}, storing and searching for pre-visited regions \cite{collins2020efficient}, and finding the global optimal path from distinctive topologies \cite{rosmann2017integrated}. Our work also evaluates paths from distinctive topologies, though we focus on path planning with consideration about the path's perception quality.  

To create a traversable topological graph structure from a given environment, PRM-based methods and generalized Voronoi diagram (GVD) can be used. However, as pointed out in \cite{rosmann2017integrated}, GVD is more beneficial than PRM-based methods in global planning since GVD guarantees coverage of the environment. GVD is a form of a roadmap structure, which can be computed online via the Euclidean signed distance field (ESDF) with low computational cost \cite{lau2013efficient}. Although topology in 3D spaces can be computed to better describe MAV flight as in \cite{oleynikova2018sparse}, in this study, we use a 2D topological graph for the sake of efficient computation under the assumption that the flight altitude would not drastically change. 

\vspace{-1mm}

\begin{figure}[tb]
  \centering
  \includegraphics[width=0.95\linewidth]{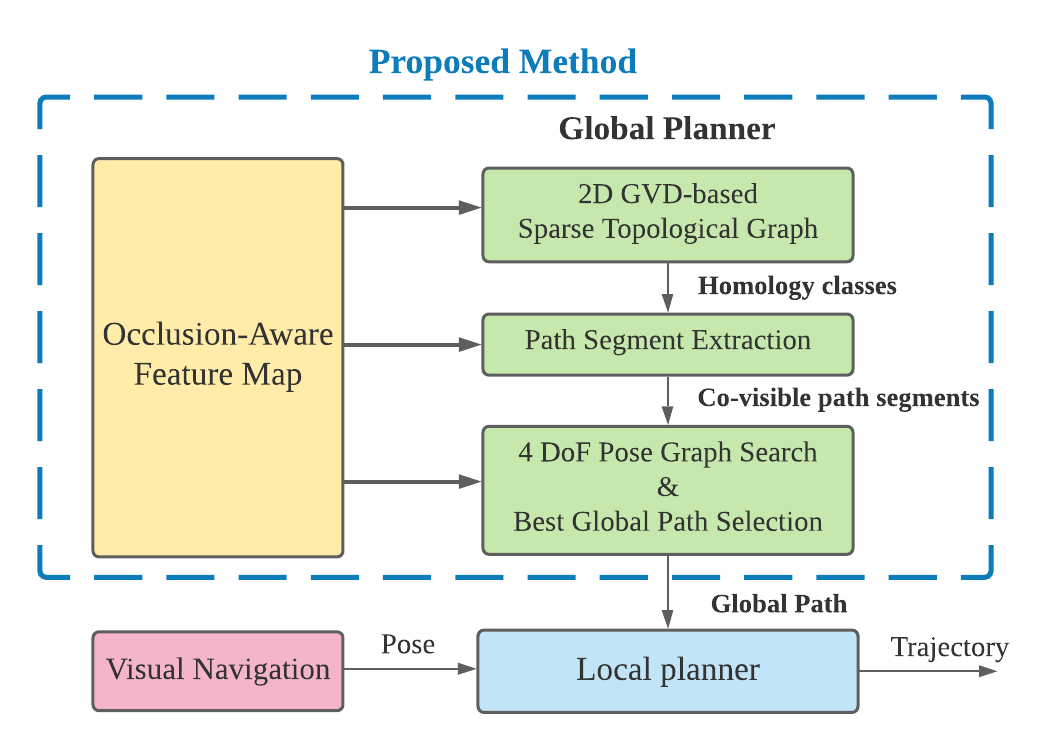}
  \caption{Overview of the proposed method.} 
  \label{figure_overview}
  \vspace{-1mm}
\end{figure}
\section{SYSTEM DESCRIPTION}

In this paper, we suggest a planner that allows MAV to move from the start to  the goal point while maintaining good self-localization. 
% The perception system of MAV obtains image about the surrounding environment through monocular camera mounted on the body.
The MAV can observe surrounding environments and estimate ego-motion using a vision sensor. Although proposed method can also be adapted to multi-camera systems, but single camera is used for demonstration.

The overall structure for the perception-aware motion planning is described in Fig. \ref{figure_overview}. To represent the environment, we used an occlusion-aware feature map, which is an integrated map of volumetric and landmark maps. This allows to obtain information about visibility of map points and occupancy of the environment, which is required for evaluating the perception quality and generating a collision-free path. The global planner first constructs a sparse topological graph via GVD. Then initial paths belonging to distinct topologies are extracted from the graph. For each initial path, it is divided into multiple segments based on visibility information.  Pose samples are generated near segments, and perception quality of each sample is evaluated. Then the path  which can obtain the maximum perception information is generated via graph search. Among the generated paths from distinctive topologies, the best path with respect to the path length and perception quality is selected as an output of the global planner.
% After global planning, a local planning module creates a smooth and feasible trajectory based on the initial global path and current estimated pose from visual navigation system.

\section{MAP REPRESENTATION} \label{map representation}
\begin{figure}[tb]
\centering
\includegraphics[width=0.99\linewidth]{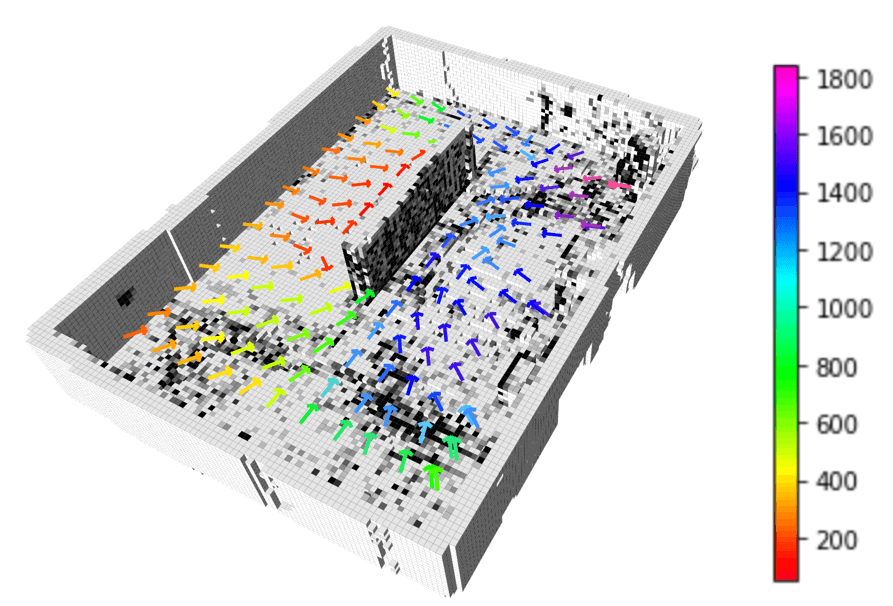}
%\\
%\includegraphics[width=0.7\linewidth]{figure/topology_difference2.png}
\caption{Occlusion-aware feature map representation of the environment drawn in Fig. \ref{figure_topology_difference}. The darker the voxel color is, the more landmark information it contains. The arrows in the figure indicate the orientation with maximum visible landmarks from sampled position, and the color of each arrow represents the number of visible landmarks from the respective pose.}
\label{ex_map_representation}
\end{figure}

 In perception-aware planning, it is required to compute visibility information of landmarks from arbitrary poses to evaluate perception quality at candidate waypoints. In this paper, we consider visual SLAM algorithms which represent the map using point features observed in keyframes \cite{mur2017orb}, \cite{forster2014svo}. While sparse pointcloud map allows efficient query of visible landmark candidates using adjacent keyframes, it does not provide the capability to consider geometry of the scene and exclude occluded landmarks. This may limit the accuracy of the queried visibility information, especially in an obstacle-filled environment where occlusion of landmarks occurs frequently. To tackle this issue, we have considered 3D volumetric map representation which allows more accurate reasoning on 3D geometry of the scene.
 
3D volumetric map representations have been developed to enable robots to differentiate between the traversable and occupied spaces. By modeling 3D space with probabilistic occupancy grid \cite{hornung2013octomap} or ESDF \cite{oleynikova2017voxblox}, these map representations inherently provide the ability to reason about scene geometry. We devise a method to integrate SLAM map with volumetric map representation to create occlusion-aware feature map representation. Our method is similar to the method proposed in \cite{costante2018exploiting}, where the authors stored texture information for each occupied voxel's surface. In our case, instead of storing photometric information, we embed each landmark's information in the occupied voxel corresponding to its location. Example of this map representation is illustrated in Fig. \ref{ex_map_representation}. 

To construct the integrated map of an environment, we first construct a volumetric map and a SLAM map separately from series of measurements by RGB-D camera. Then the respective global coordinate frames of the maps are aligned  and each landmark's information is embedded into the voxel at its location. Finally, to allow raycasting-based visibility query, landmark information is shifted to the nearest surface voxel which is visible from reference keyframe's pose. In this work, we used ORB-SLAM2 \cite{mur2017orb} and Voxblox \cite{oleynikova2017voxblox} for SLAM map and volumetric map representation, respectively.

\section{TOPOLOGICAL GLOBAL PLANNING}

\begin{figure}[tb]
    \centering
  \subfloat[]      
  {\includegraphics[width=0.47\linewidth]{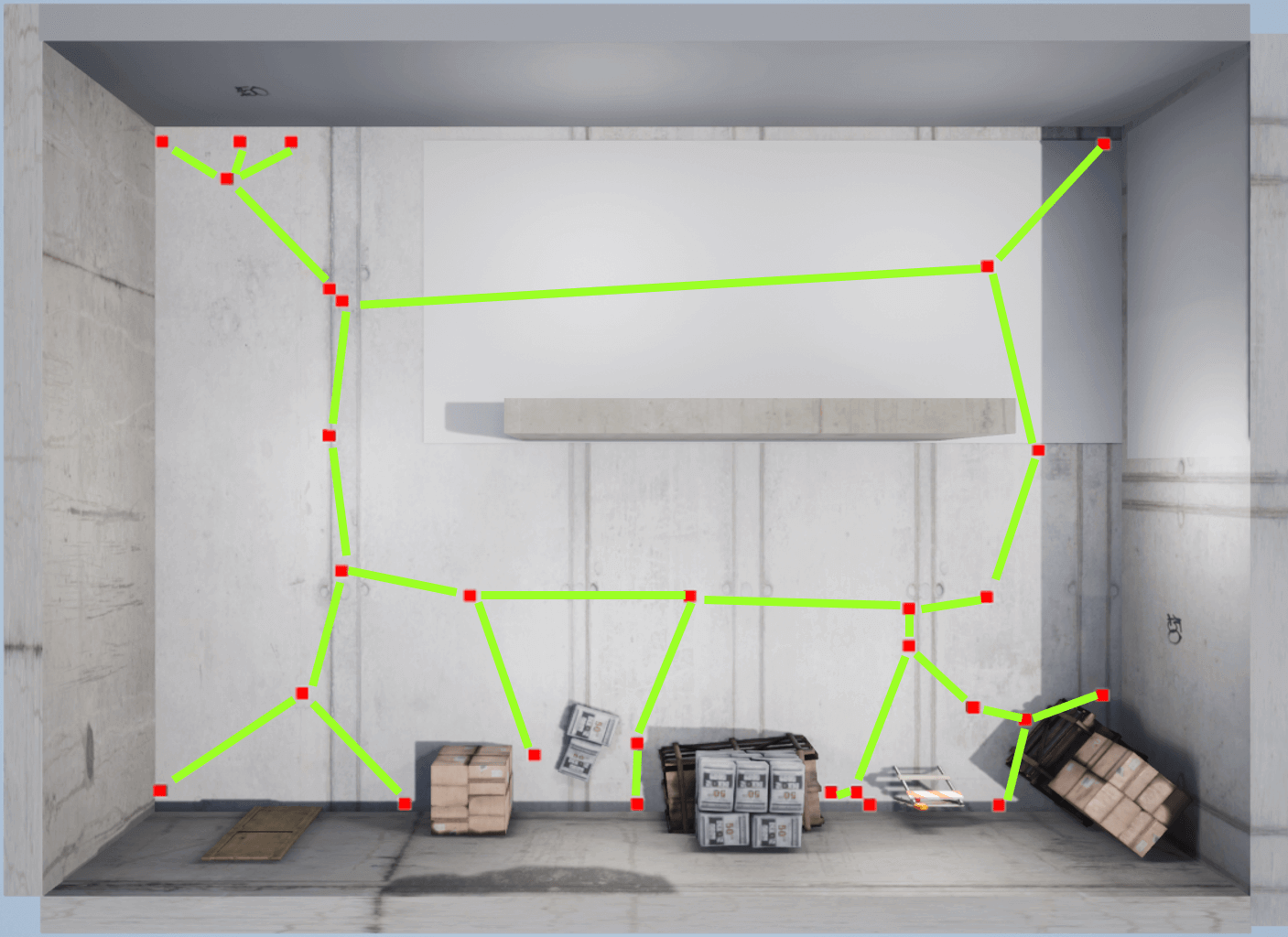}} 
  \subfloat[]   
 { \includegraphics[width=0.47\linewidth]{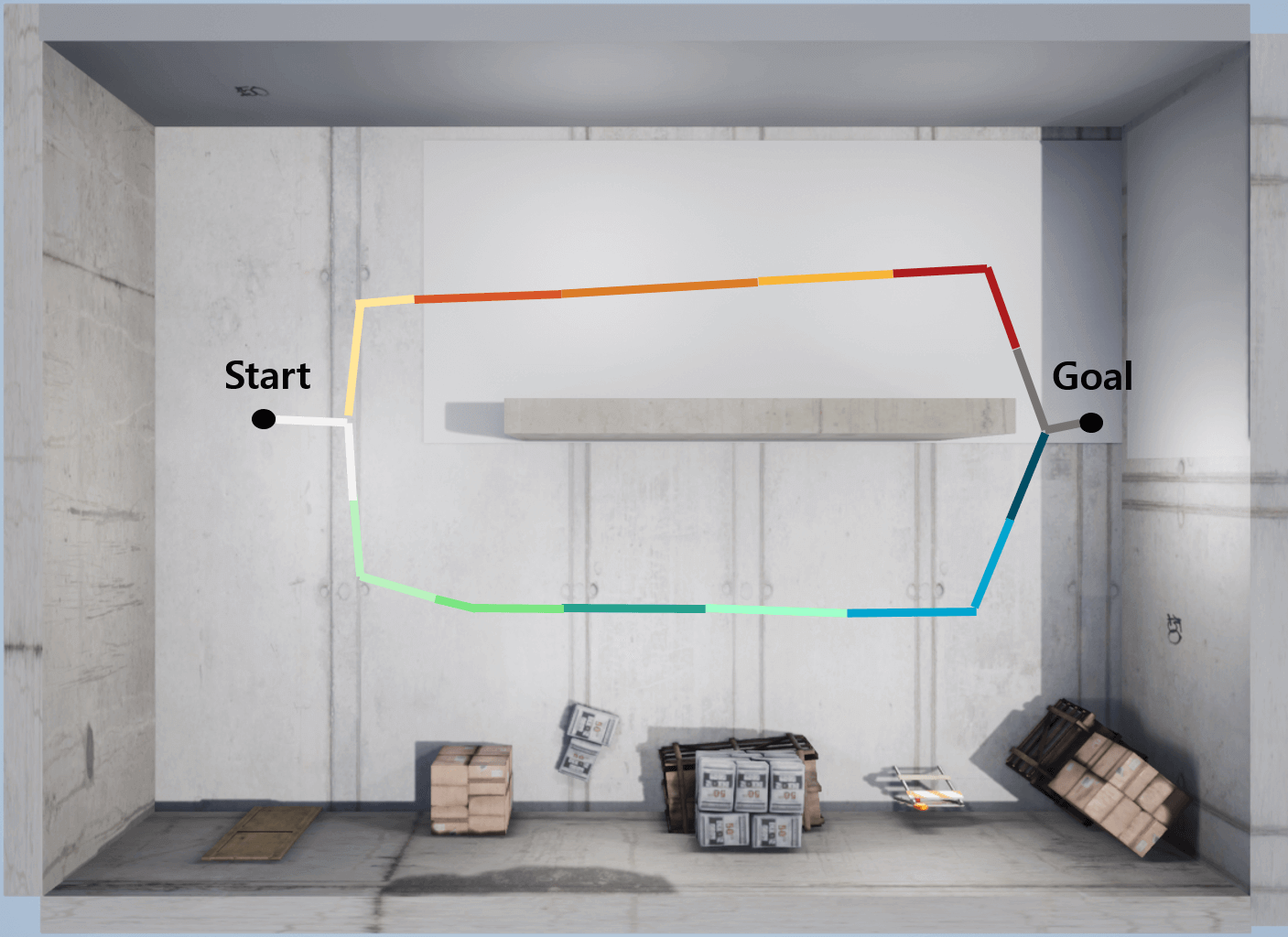}}  \\[-2ex]
  \subfloat[]      
  {\includegraphics[width=0.47\linewidth]{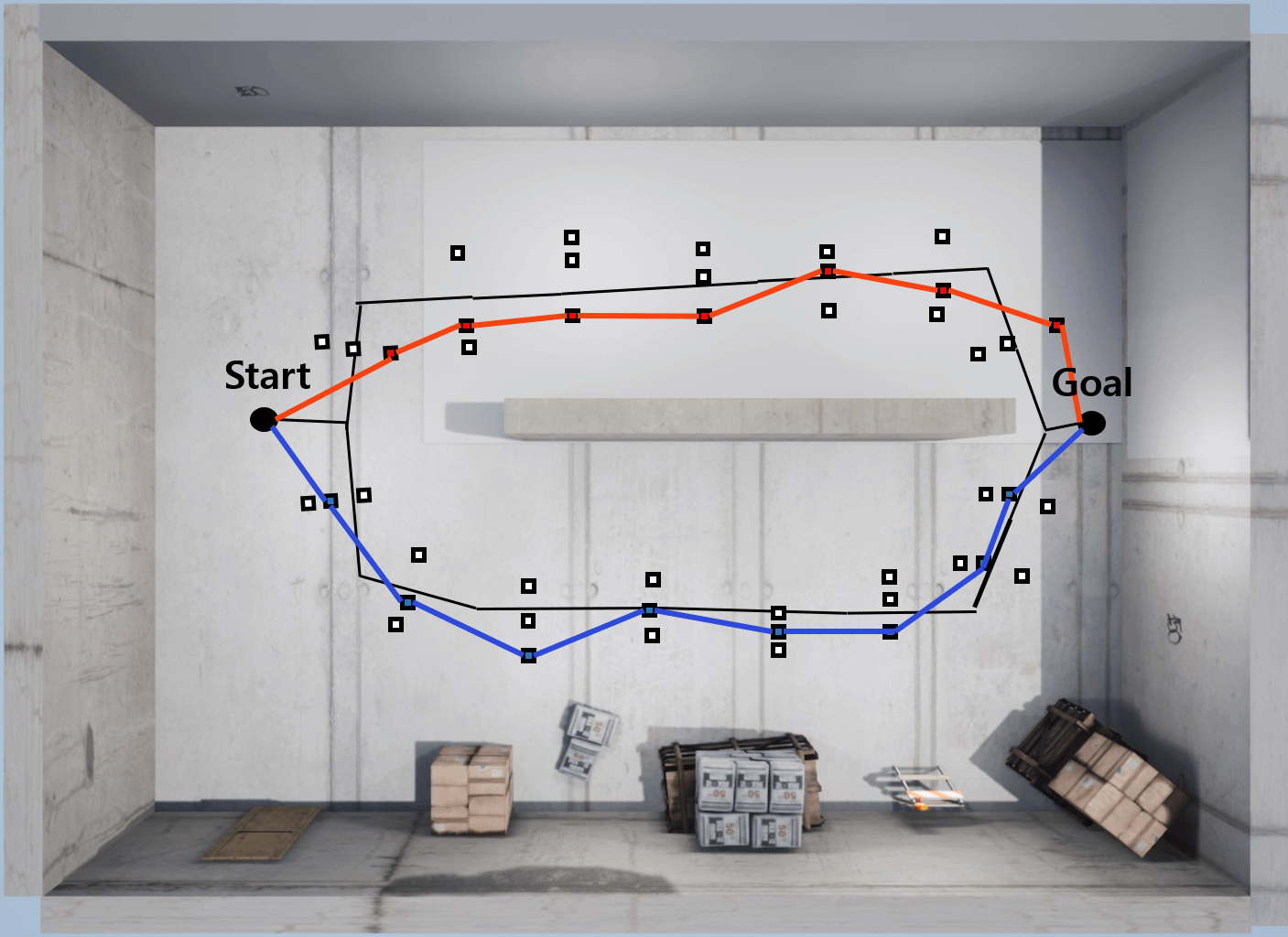}} 
  \subfloat[]  
 { \includegraphics[width=0.47\linewidth]{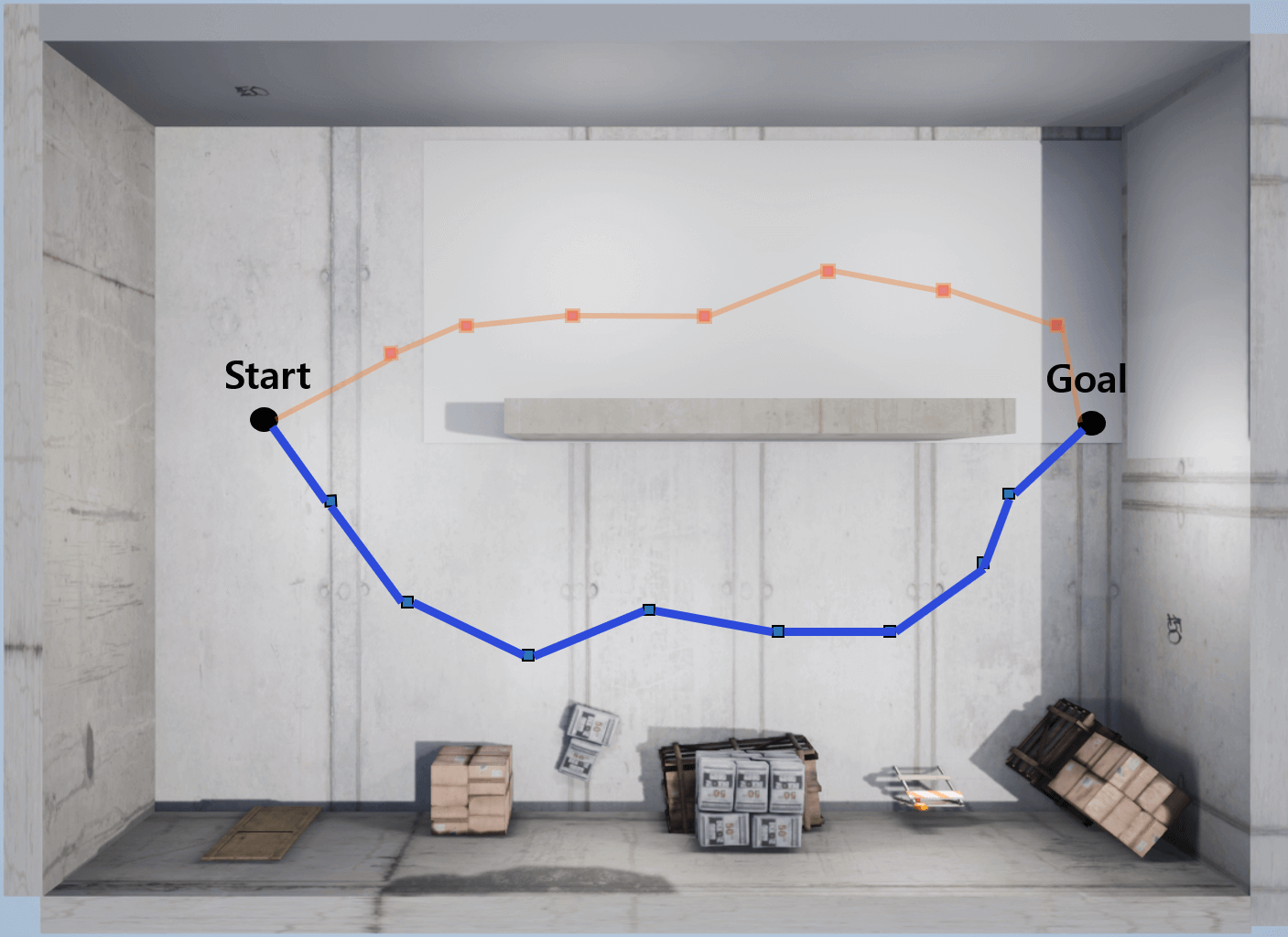}} 
    \caption{Process of the proposed planner (a) Topological graph generation from GVD (b) For each homology class, path segments are extracted based on feature co-visibility. (c) Pose samples are generated and optimal paths are found. (d) The best path is selected with respect to the perception quality and path length. }
    \label{figure_process}
    \vspace{-2mm}
\end{figure}

Based on the integrated map, we generate a global path which serves as an initial reference for a low-level planner.  
%\ys{From the observation that visual information and relative position between MAV and obstacles are strongly related,}{이게 진짜 중요한 intuition이자 동시에 contribution이라서 문장 늘려서 풀어서 자세하고 강하게 설명해주면 좋을거 같애! 로봇의 로컬 플래너는 initial path을 기준으로 순간순간 상황에 맞게 수정하여 경로를 만들기 때문에, 초기 경로 + 해당 초기 경로에서 조금 detour 경로도 퍼셉션이 실패하지 않아야 함. 토폴로지 레벨에서 경로를 결정하는건 블라블라 때문에 이러한 요구 조건을 굉장히 잘 반영하고 좋은 초기 경로를 선택할 수 있게 해준다. 그래서 우리는 블라블라 하는 방법론을 제안한다.} These family of trajectories usually share topological equivalence, 
From the observation that visual information and relative position between MAV and obstacles are strongly related, we devise a method to generate multiple global paths with distinctive topologies and to select the path with respect to the perception quality and path length. This section is configured as follows. In Sec. \ref{path_homo_perc_qual}, we will give a detailed explanation on why it would be beneficial for a perception-aware planner to consider topological properties of the path. The process of perception-aware planning will be covered in Sec. \ref{topo_graph_generation} $\sim$ \ref{selection}.
The overall process of global planner is illustrated in Fig.  \ref{figure_process} and Alg. \ref{alg: Planner}.
\subsection{Homology Classes and Perception Quality}\label{path_homo_perc_qual}

\begin{figure}[tb]
\centering
\includegraphics[width=0.9\linewidth]{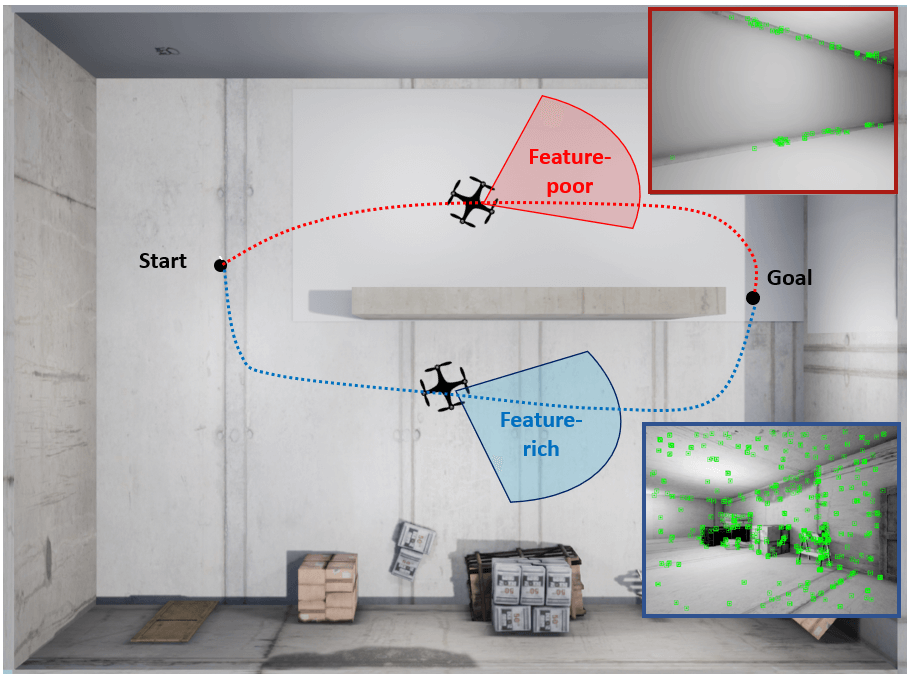}
\caption{ Illustration of dependence of perception quality on homology classes. Tracking a trajectory belonging to homology class marked in red leads to bad localization because feature-rich surfaces are occluded by the wall in the middle. Therefore, homology class of blue path is preferred in terms of perception quality. }
\label{figure_topology_difference}
\end{figure}

%\ys{Roughly speaking,}{빼는게 좋을거 같애! 본 연구에서, Homology class는 ~~으로 정의한다.로 이 문장 구성할까??} 
In this subsection, the close relationship between topological planning and perception quality is explained. To express topological equivalence of trajectories, the concept of homology class is widely used. Two trajectories with fixed start and goal points in 2D belong to the same homology class if the cycle formed by them does not include or intersect any obstacle. For a more formal definition of homology class, refer to \cite{bhattacharya2012topological}.

During trajectory generation for MAV flight, the local planner refines the reference path from global planner to smooth and feasible trajectory. In order to maintain visual navigation stable, the global planner should generate a path such that navigation does not fail, not only in a reference path but also in a path refined by the low-level planner. 
However, the local planner module updates the path with collision avoidance constraint. For example, some algorithms restrict the search space into free, convex region as a hard constraint \cite{richter2016polynomial}, \cite{park2020efficient} and other algorithms which use gradient-based optimization update paths to the opposite direction of obstacles \cite{zucker2013chomp}, \cite{zhou2019robust}. Therefore, it is difficult for  local planners to update a path to jump over obstacles and change its homology class, thus reference and refined paths are topologically equivalent. 

On the other hand, a path's homology class affects visual information which MAV can obtain by following the path. Selecting the homology class fixes relative topology of the path to the obstacles, which determines visible surfaces of obstacles. In the vision-based navigation system, features are detected on the surface of obstacles. If feature-rich surfaces are hindered by occlusion, we can conclude that corresponding homology class is disadvantageous for visual navigation. Thus, the homology class which can guarantee visibility of feature-rich surfaces is preferred. The relationship between homology class and perception quality of the path can be observed in Fig. \ref{figure_topology_difference}. 

Therefore, searching distinctive homology classes can be a helpful heuristic for finding a reference path for reliable visual navigation. In addition, it also enables to boost computation by searching each homology class in parallel, exploiting multi-process CPU. 
%In addition, global path's homology class is maintained after the local trajectory optimization. 

\subsection{Topological Graph Generation} \label{topo_graph_generation}

\begin{algorithm}[tb]
\caption{Topological perception-aware path planner} \label{alg: Planner}
\begin{algorithmic}[1]
\STATE \textbf{Input}: Map $M$, Start $\mathbf{s}$, Goal $\mathbf{g}$
\STATE \textbf{Output}: Global Path $P^{*}$ 
% \FOR{$k$=1 to $K$} 
\STATE Topological Graph Generation $\mathcal{G} = \mathcal{(V,E)}$  
\STATE Generate Initial Paths from Distinct Homology Classes $H=\{h_1, \cdots h_T\}$
\FOR{$h_t \in \{h_1 \cdots h_T$\}}
\STATE Extract Path Segments \\
       $h_t \rightarrow s(\mathbf{s, p_1}), s(\mathbf{p_1, p_2}), \cdots , s(\mathbf{p_m, g})$
\STATE Generate 4DoF pose samples ($n_k^j$) \& evaluate perception quality $I(n_k^j)$
\STATE Pose graph search $P^{*}_t=(n_{k_1}^1, n_{k_2}^2, \cdots, n_{k_N}^{N})$ 
\STATE Evaulate quality of the path $ q(P_t^{*})$ \eqref{path_cost}
\ENDFOR 
\STATE $P^{*} = \argmax\limits_{q(P_t^{*})}{ \{P_1^{*}, \cdots P_T^{*}\}} $
% \STATE \begin{align*}
% \end{align*}
\end{algorithmic}
\end{algorithm}

\begin{figure}[tb]
\centering
\includegraphics[scale=0.47]{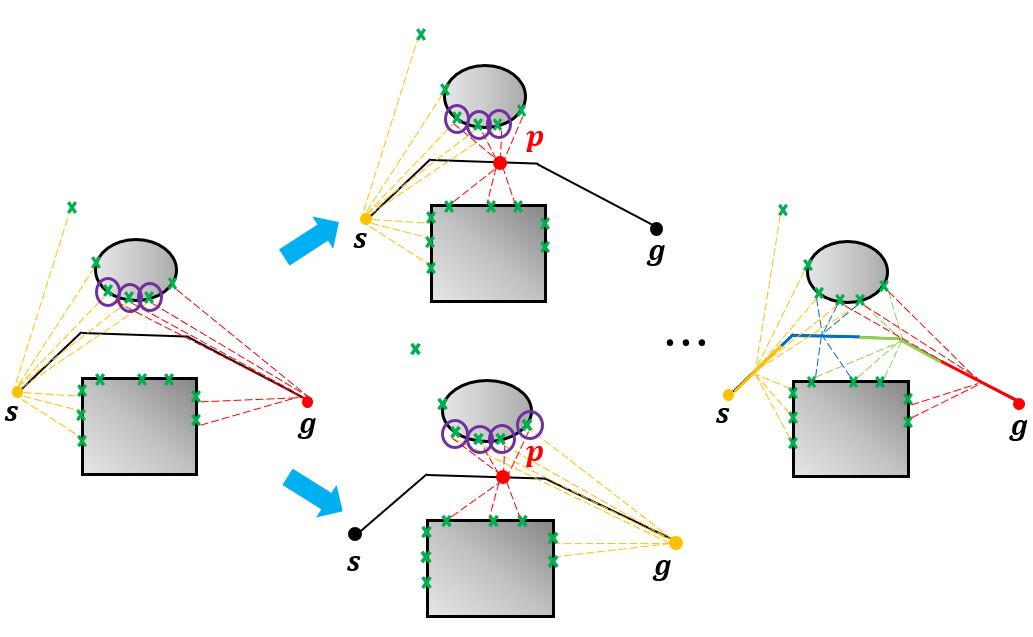}
\caption{Illustration of path segment extraction process. Green cross represents landmarks, and the co-visible candidate set is marked in purple circles. A path is iteratively divided until there are significant portion of co-visible landmarks at both ends of every segment.}
\label{figure_path_segment}
\vspace{-2mm}
\end{figure}

The methodology that we present aims to find a global path through the topological structure of the environment. To this end, it is necessary to generate topological graphs in a given environment and extract homology classes. To generate a topological graph, we create a 2D GVD with the given reference height. Construction of GVD follows the method suggested by \cite{lau2013efficient}, in which the ESDF map is used to obtain a 2D voxel map of GVD. Since the graph structure of GVD is needed, vertices and edges are extracted from the GVD. A voxel is a vertex if 1 or more than 2 neighboring voxels are elements of GVD. And a connected set of voxels is an edge if it connects two vertices. By classifying each voxel as a vertex or an edge, a bi-directional topological graph is acquired. By adding the start and goal points to the graph, connected paths from the start to the goal can be obtained from this graph. By adopting the methodology from \cite{rosmann2017integrated}, we can find `initial paths', which are sets of consecutive vertices, from distinct homology classes using breadth first search on the topological graph. 
%and H-signature as a topological invariant to check redundant homology classes.
%\ys{For detailed descriptions, please refer to \cite{rosmann2017integrated}.}{요거 빼도 될듯! 아니면, please refer to 부분만이라도 빼야할거 같애!}

\subsection{Extracting Path Segments} \label{subsec_path_seg}
After initial paths are generated in different homology classes, each path and corresponding homology class are evaluated with respect to perception quality. It is important to maintain co-visibility to the landmarks along the path for vision-based localization. Thus, we evaluate the perception quality of the path based on co-visible landmarks while traveling along the path. However, especially in environments with multiple obstacles, visible landmarks change as the robot travels through the path, which makes it difficult to find a set of globally co-visible landmarks along the whole path.

Thus, we seek to split the initial path into smaller `path segments' where points in the same segment share a significant amount of visible landmarks. This is inspired by the previous study on co-visibility based regional segmentation \cite{blanco2006consistent}, where different positions in the map are grouped based on the similarity of their observations. While the method in \cite{blanco2006consistent} was devised to obtain a topological map of the environment, our objective is to divide a path based on the similarity of predicted observation along the path. The algorithm is depicted in Fig. \ref{figure_path_segment}. 

We defined the `visibility candidate set' at point $\mathbf{p}\in \R^3$, $V(\mathbf{p})$, as the set of the landmarks which are within the vertical FoV and sensing range.
Given two points $\mathbf{p, q}\in \R^3$, the `co-visible candidate set' $CV(\mathbf{p, q})$ and `co-visibility ratio' $CR(\mathbf{p, q})$ are respectively defined as
\begin{align}
    &CV(\mathbf{p, q}) = V(\mathbf{p}) \cap V(\mathbf{q}) \\
    &CR(\mathbf{p, q}) = \frac{|CV(\mathbf{p, q})|}{|V(\mathbf{p}) \cup V(\mathbf{q})|}
\end{align}

%As in Fig. \ref{figure_path_segment}., from two consecutive vertices $(v_i, v_{i+1})$ in the initial path, we first evaluate $CR(v_i, v_{i+1})$. If it is larger than the threshold $\eta \in [0,1]$, we consider the line between vertices as a path segment and assign $CV(v_i, v_{i+1})$ as the co-visible candidate set for the path segment. Otherwise, we query the visibility candidate set at the midpoint $p$ of $(v_i, v_{i+1})$ and evaluate $CR(v_i, p)$ and $CR(p, v_{i+1})$. We iteratively divide the path until the co-visibility ratio of the end points for each segment becomes larger than $\eta$, or  length of the segment reaches the minimum length $\ell_{min}$. As a result, an initial path $h$ is transformed into multiple path segments $(s_1, s_2, \cdots, s_m)$ where the points in the same segment share significant co-visible landmarks. This allows to approximate the perception quality of initial path with finite number of path segments and their co-visible candidate set. \textcolor{purple}{위에 This allows ... 랑 아래 문장 다시 보니까 중복되는거 같음.}
%Co-visible candidate set represents trackable landmarks all over the path segment. 
As in Fig.  \ref{figure_path_segment}, we begin with evaluating co-visibility ratio of the start and goal point: $CR(\mathbf{s}, \mathbf{g})$. If it is larger than the threshold $\eta \in [0,1]$, we consider the path as a path segment $s(\mathbf{s, g})$ and assign $CV(\mathbf{s, g})$ as the co-visible candidate set for the path segment. Otherwise, we query the visibility candidate set at the midpoint $\mathbf{p}$ of the path and evaluate $CR(\mathbf{s, p})$ and $CR(\mathbf{p, g})$. We iteratively divide the path until the co-visibility ratio of the end points for each segment becomes larger than $\eta$, or  length of the segment reaches the minimum length $\ell_{min}$. As a result, an initial path is transformed into multiple path segments $s(\mathbf{s, p_1}), s(\mathbf{p_1, p_2}), \cdots , s(\mathbf{p_m, g})$, where the points in the same segment share significant co-visible landmarks. Furthermore, the previously stored co-visible candidate set of the path segment is used during the evaluation of perception quality to avoid redundant querying of visible landmarks, which include time-consuming raycasting.

\subsection{Pose Graph Construction \& Graph Search}

While a path over sparse graph can represent the homology class, directly using it as global path precludes the existence of a path with better perception quality within the homology class. 
%Also we should consider limited FoV of the camera to check  visibility of the landmarks at given pose. 
Also, since GVD is extracted from the 2D plane at the reference height, 3D obstacles cannot be considered in the sparse graph. From this need, we construct a dense graph from given sequence of path segments as in Fig. \ref{figure_process} (c). Each node of the dense graph represents a 4 Degree of Freedom (DoF) pose including the position and yaw angle of the MAV ($x, y, z,$ and $\theta$), and graph search is performed to find the optimal 4 DoF path.

The 4 DoF pose graph is constructed as follows. First we divide the initial path  into intervals of length $\ell$  which is determined by nominal speed $v_{nom}$ and time step $Ts$. 
Each of these intervals represents a \textit{layer} of the graph; only the nodes within consecutive layers are allowed to form edges. 
We denote layers as $L^1, L^2, \cdots L^N$ where $N$ is the total number of the intervals obtained by splitting the entire path. 
% For each layer $L^j$, waypoint candidates are randomly sampled within a distance $R_{sample}$ centered around the path, \textcolor{red}{on the plane passing through the starting point of the interval and perpendicular to the edge of the interval}.
For each layer $L^j$, waypoint candidates are randomly sampled from the plane passing through the starting point of the interval and perpendicular to the edge of the interval. 
Samples are generated within a distance $R_{sample}$ centered around the path. 
On top of this, yaw angles are sampled at equal intervals. The sampled waypoint candidates and yaw angles are combined to form nodes of the layer $\{n^{j}_1, \cdots n^{j}_m\}$, where $n^{j}_k=(x^j_k, y^j_k, z^j_k, \psi^j_k)$. 
Nodes in the consecutive layers are connected only if the difference of yaw angles between the nodes is smaller than a certain limit $\dot{\psi}_{lim}$ to prevent an abrupt yaw change.

To find the path with maximum visual information, each node's perception quality needs to be evaluated. We use the Fisher information matrix (FIM) as the metric for perception quality, which quantifies the information that can be obtained about the desired state through measurement. By modeling the camera as a bearing sensor as in \cite{zhang2020fisher}, FIM of measurement on a landmark located at $\mathbf{l}$ observed from pose $\mathbf{x}$ can be formulated as
\begin{gather}
\mathrm{FIM}(\mathbf{l;x})= \frac{1}{\sigma^2} (\mathbf{J(\mathbf{l}; x)})^T \mathbf{J(\mathbf{l}; x)}\\
\mathbf{J(\mathbf{l}; x)}=\Big(\frac{1}{||\mathbf{l}^c||}\mathcal{I}_3 - \frac{1}{||\mathbf{l}^c||^3}\mathbf{l}^c(\mathbf{l}^c)^T \Big) \begin{bmatrix}\mathcal{I}_3 & [\mathbf{l}^w]_\times \end{bmatrix}
\end{gather}
where $\mathbf{l}^c, \mathbf{l}^w$ are the position of point seen from camera frame and global frame respectively, and $\sigma$ is standard deviation of measurement noise. For each node, we compute the visible landmarks from previously stored co-visible candidate set along the path segment containing the layer node is in. For the $j$-th node, we evaluate the FIM of the visible landmarks at pose $n^{j}_{k_{j}}$ and denote it as $\mathrm{I}(n^j_{k_j})$.

We perform a graph search over the pose graph to find the 4 DoF path with the smallest value of combined distance cost and perception cost. Given a path $\mathrm{P}$ as a sequence of connected nodes $\mathrm{P}=(n^1_{k_1}, n^2_{k_2}, \cdots ,n^N_{k_N})$, we can formulate the graph search problem as 
\begin{align}
   \mathrm{P}^*=\argmin_{k_1, k_2, \cdots, k_N} \lambda_d c_d(\mathrm{P}) - \lambda_p c_p(\mathrm{P}),
\end{align}
where $\lambda_d$, $\lambda_p \in \R$ are weights for distance cost and perception cost, the distance cost $c_d(\mathrm{P})$ can be defined straightforward as the length of total path.  The perception cost of the path $c_p(\mathrm{P})$ is formulated as
\begin{gather}
c_p( \mathrm{P}) = \frac{1}{N} \sum_{i = 1}^{N} \log (\mathrm{det}(\mathrm{I}(n^i_{k_i})).
\end{gather}
% \colorcomment{Dabin: Mention about DAG and Topological Sorting since this assures the time complexity.}

To perform graph search, a layered structure of the graph can be exploited. Since each layer is arranged in time order, the graph is a directed acyclic graph. Also, topological sorting is used for graph search by dynamic programming. It can find an optimal path 
with less computation than Dijkstra search \cite{cormen2009introduction}.
% time complexity of $O(|V|+|E|)$, where $|V|,|E|$ are number of total vertices and edges respectively.

\subsection{Selection of the best path}\label{selection}

We design the cost function to quantify the quality of the generated path $P^{*}$ for each homology class as
\begin{align}\label{path_cost}
&q(\mathrm{P^{*}}) = \eta_d \big( \frac{d}{d_{min}}-1 \big) - \eta_p f(\mathrm{c}_{p, min} - {\mathrm{c}_{p, thr}}), \\
&\mathrm{c}_{p, min} = \min( c_p(s(\mathbf{s, p_1})), \cdots, c_p(s(\mathbf{p_m, g}))) \label{path_segment}
\end{align}
 where $f$ is defined as $f(x) = 1/(1+\mathrm{exp}(x))$, $d$ is the distance of the path, $d_{min}$ is the length of the shortest path among the generated path candidates, and $\mathrm{c}_{p,thr}$ is the parameter to indicate the required information for robust visual navigation. As in \eqref{path_segment}, the perception quality of the path is determined by the minimum $c_p$ value among the segments of the path. It enables to avoid selecting the path passes through feature-poor region, which would result in high estimation error. The reason for using the sigmoid function is because effect of FIM on the localization performance degrades if there is enough information.
 Among the selected best path within each homology class, the path with the smallest cost is selected as the global path.

 \begin{table}[tb] 
\footnotesize{
   \caption{\small{Parameter List for Simulation}}
    \centering
    \begin{tabular}{|m{1.4cm}|m{3.6cm}|m{2.0cm}|}
    \hline
        Types & Parameter Name & Value \\
        \specialrule{.10em}{.10em}{.10em}  
        Segment & Min Segment Length [$m$] & $l_{min} = 0.5$ \\
        %  \cline{2-3}
         Extraction & Covisibility Ratio  & $\eta = 0.5$ \\
         \hline
         \multirow{7}{*}{Pose Graph} & Nominal speed [$m/s$]  & $v_{nom}=0.4$  \\
        %  \cline{2-3}
         & Time Step [$s$] & $Ts=1.0$  \\ 
        %  \cline{2-3}
         & The number of Samples & $10$   \\
        %  \cline{2-3}
         & Max Radius [$m$]& $R_{sample}=0.4$ \\
        %  \cline{2-3}
         & Yaw Rate Limit [$rad/s$] & $\dot{\psi}_{lim}=0.3$\\
        %  \cline{2-3}
         & Edge Distance Weight & $\lambda_d=0.1$\\
        %  \cline{2-3}
         & Edge Perception Weight & $\lambda_p=1.0$\\
         \hline
         \multirow{3}{*}{Selection} & Perception Quality Threshold& $c_{p, thr}=6.0$ \\
        %  \cline{2-3}
         & Path Distance Weight & $\eta_d = 0.2$ \\
        %  \cline{2-3}
         & Path Perception Weight & $\eta_p = 1.5$ \\
         \hline
    \end{tabular}
    \label{tab:parameter}
    }
\end{table}

\vspace{-2mm}
\section{VALIDATION}

To validate the proposed topological global planner, simulations and experiments were performed. The global planner was connected to local trajectory optimization to obtain a smooth and kinodynamically feasible trajectory. We used the gradient-based optimization method proposed in \cite{zhou2019robust} as an example of local planner. 
% For creating feasible path with fast computation, position and yaw were optimized respectively.\colorcomment{Isn't this a repetition of methodology part? Or at least it sounds like. Alternative : Both position and yaw angle trajectory were optimized through B-spline optimization respectively.} 
%The path quality of the path generated by the planner is validated by implementing it on simulator and actual MAV. \textcolor{blue}{너무 개괄적인 문장인 거 같음. 또 앞문장이랑 중복이라 빼버려도 될 듯!} 
% \vspace{-2mm}

\subsection{Simulation}

\begin{figure*}[tb] 
  \centering
  \subfloat[\label{fig_path_storage}]      {\includegraphics[width=0.45\linewidth]{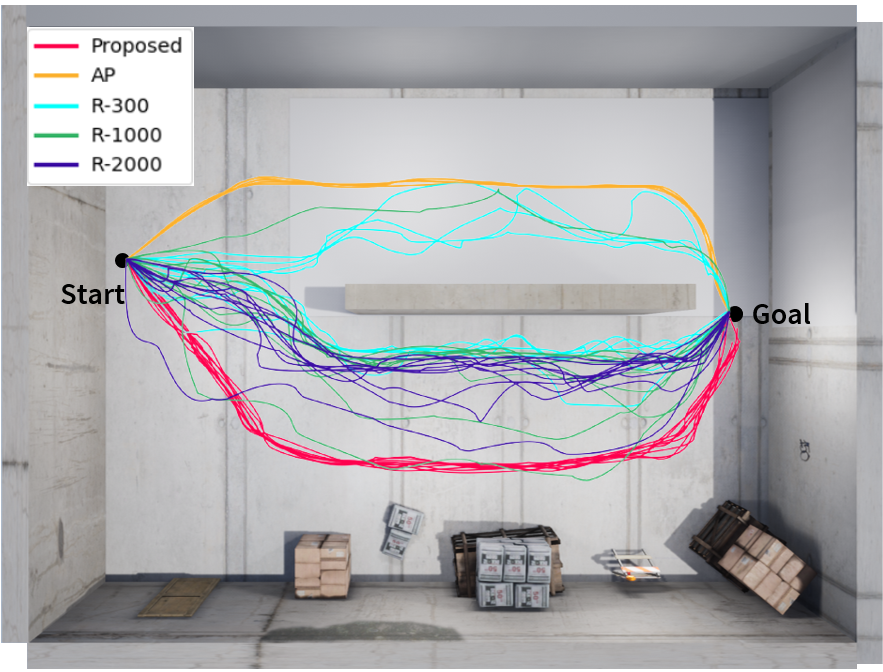}} \hspace{1.5cm}
  \subfloat[\label{fig_path_gallery}] { \includegraphics[width=0.405\linewidth]{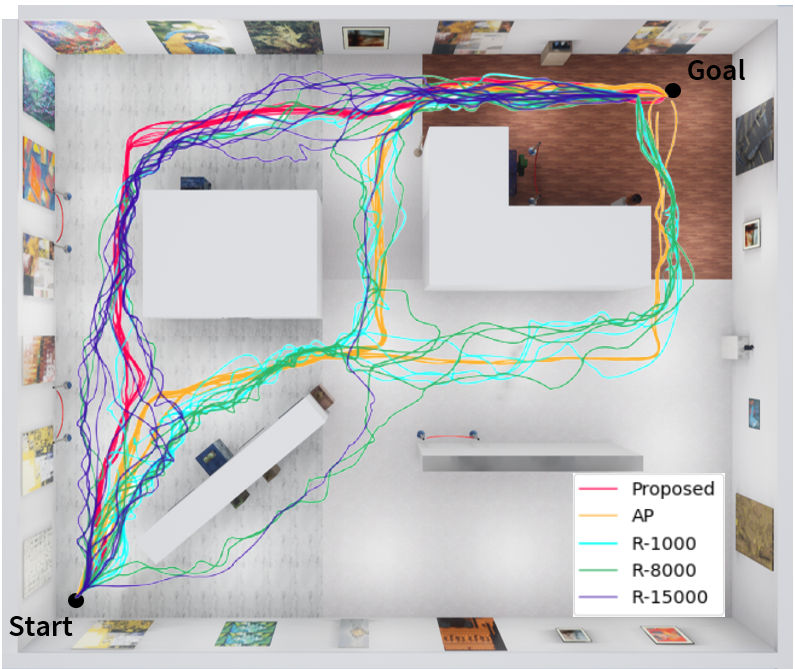}}
    \caption{ Resulting trajectories from the tested planners in (a) \textit{storage} and (b) \textit{gallery} environments. }
   \label{fig_path_figure}
   \vspace{-2mm}
\end{figure*}

We used the Unreal engine with AirSim plugin \cite{shah2018airsim} to perform high-fidelity simulations in photo-realistic environments. For visual navigation, a front-looking RGB-D camera was used and we conducted experiments in two realistic environments, \textit{storage} and \textit{gallery}, as in Fig. \ref{fig_path_figure}. Simulations were performed on a desktop with 8 core Intel i7 3.2GHz CPU and 32GB RAM. 

Four metrics were used for evaluation of the path quality. The first is the total length of the path and the second is the translational estimation error of VO algorithm, which is for quantifying the perception quality of the path. Third metric is the ratio of successful runs of VO without loss in feature tracking, which is related to the reliability of the path for stable navigation. Lastly, computation time was calculated to check computational efficiency of the algorithms. We used ORB-SLAM2 \cite{mur2017orb} as an example for VO algorithm. 

We compared the performance of the proposed method with two other methods. 1) Perception-\textbf{A}gnostic \textbf{P}lanner (AP), which %is topological planning such as the proposed method but without consideration about perception quality, 
is a topological planner but without consideration about perception quality, 2) perception-aware sampling-based planning based on RRT* similar to \cite{sadat2014feature}, with some modification to fit in our settings. In perception-aware RRT*, at each sample, visible landmarks are queried, and the perception cost is evaluated using landmarks co-visible from the sample pose and its parent's pose. The criterion for choosing the best path is comparing weighted sum of the distance cost and the perception cost, similar to the proposed planner. Three different numbers of samples were used for the sampling-based planner. Perception-aware RRT* with $N$ samples are shortened as `R-$N$'. Parameters used in the simulations are noted in Table \ref{tab:parameter}.

\begin{figure*} 
    \centering
  \subfloat[\label{fig_error_storage}]      {\includegraphics[width=0.47\linewidth]{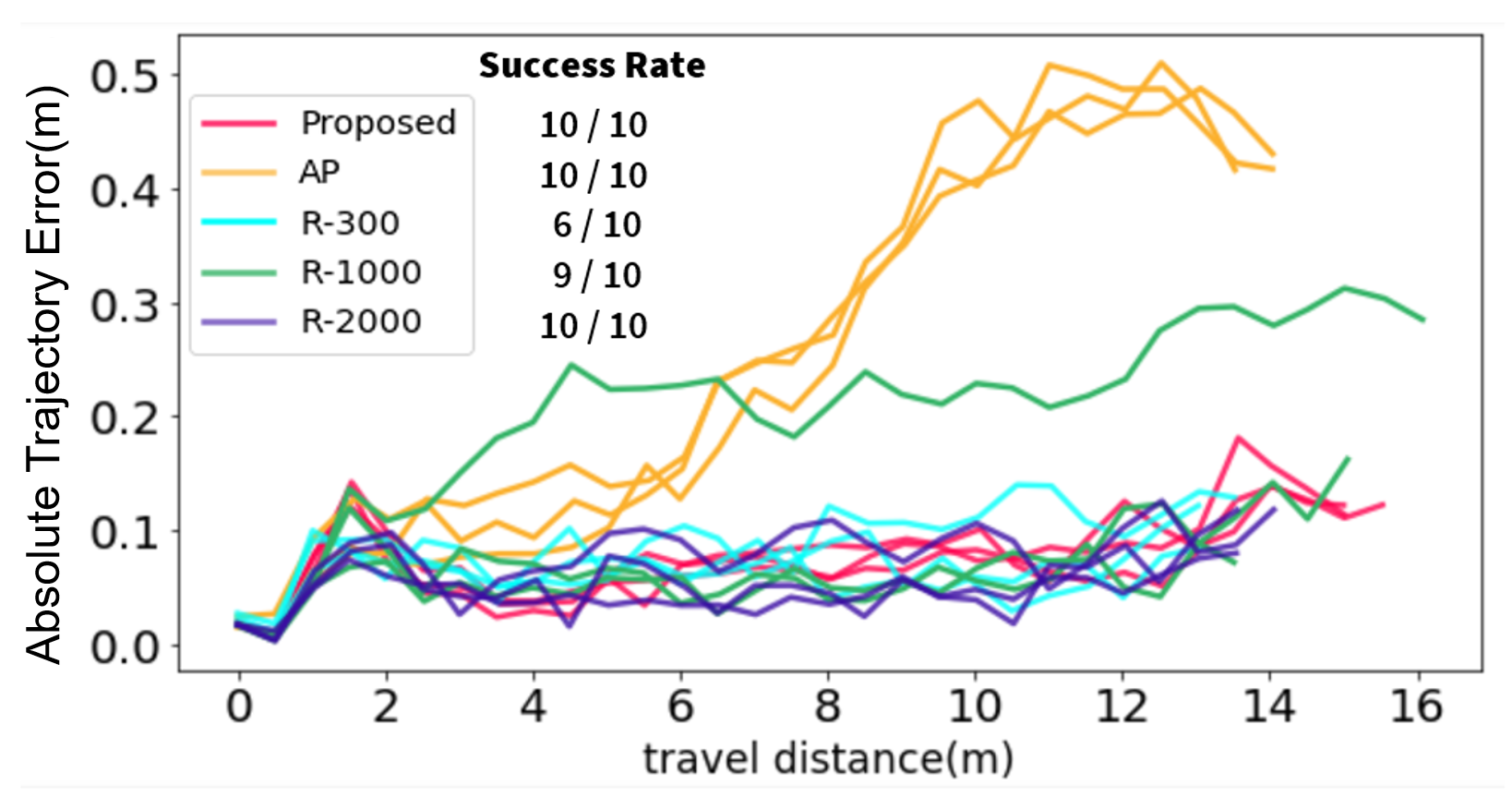}} \hfill
  \subfloat[\label{fig_error_gallery}] { \includegraphics[width=0.47\linewidth]{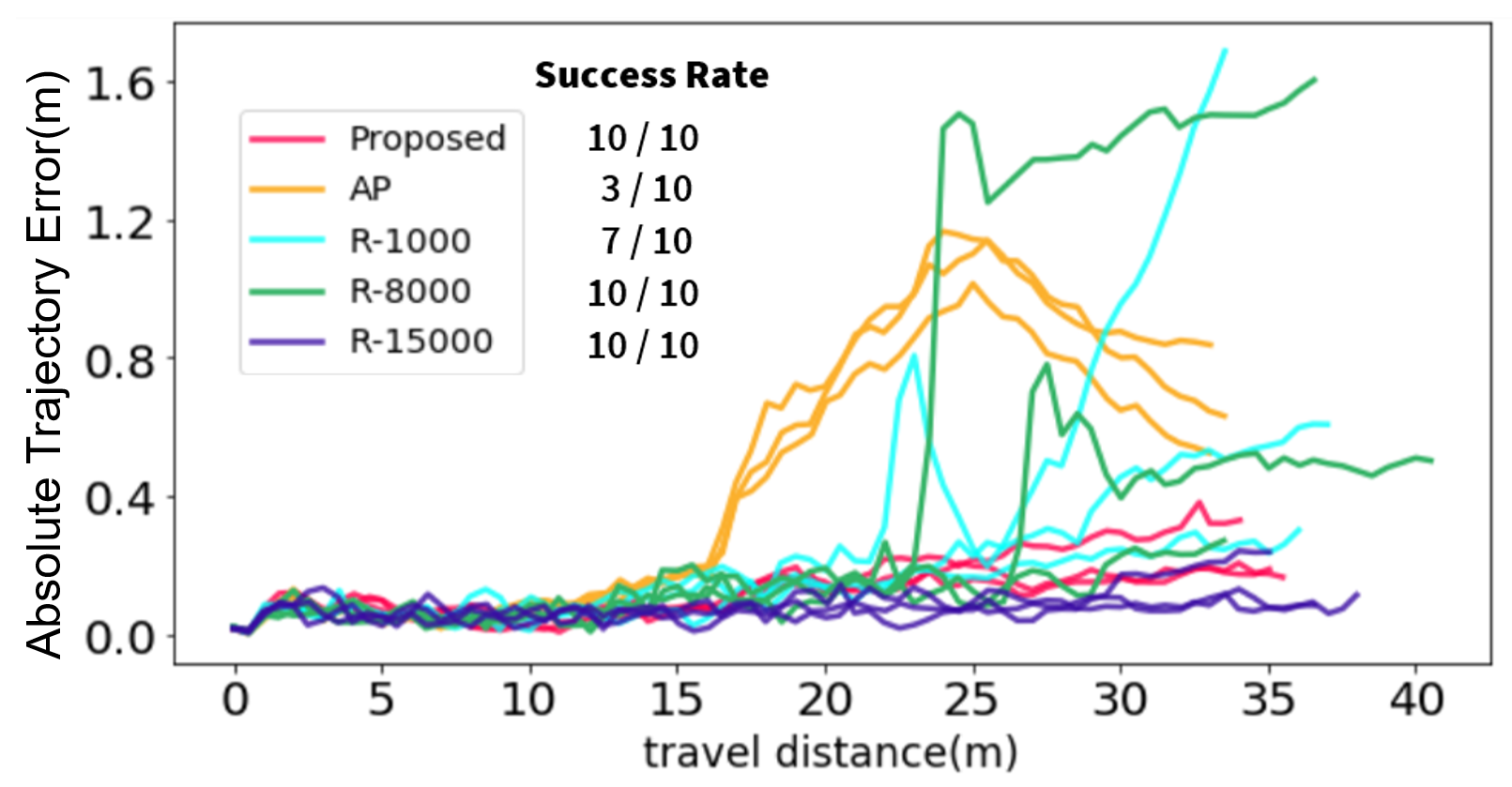}}
  \caption{  Absolute trajectory error of VO  with respect to travel distance, and success rate for each planner at (a) \textit{storage} and (b) \textit{gallery} environment. Only 3 successful trials for each planner are shown for clear visualization. }
    \label{fig_path_error}
    \vspace{-2mm}
\end{figure*}

We ran each algorithm 10 times for each scenario, and the resulting trajectories are presented on Fig. \ref{fig_path_figure}. Also, absolute trajectory error (ATE) for each path and success rates of VO are presented in Fig. \ref{fig_path_error}. For clear visualization, ATE values for only three trials are visualized for each planner.

In the first environment, \textit{storage}, there are two selectable homology classes, distinguished by the wall in the middle. The class at upper side passes through a texture-less region and the lower side has many objects with abundant visual information. As in Fig. \ref{fig_path_figure}(a), the proposed algorithm selected the lower side in every trials although path length became longer than choosing the upper side. On the contrary, AP selected the upper side’s homology class and  estimation error became higher than the proposed method's trajectory. For sampling-based planner, we chose 300, 1000, and 2000 samples respectively. Even though the sampling-based planner is designed to prefer feature-rich paths, planner with low sample numbers (300, 1000) often failed to find homology class beneficial for visual navigation and failed to maintain VO during flight. With large samples (2000), it can generate path toward feature-rich region and result in small estimation error, but it takes more computation time compared to the proposed method.

The second environment, \textit{gallery}, contains two major texture-poor regions: a horizontal corridor to the right of the center and a longitudinal corridor to the upper middle. In contrast, many texture-rich objects are distributed along the walls surrounding the environment, especially the walls of left and upper side. Similar to the previous environment, AP chose the homology class with the shortest length among 10 distinct homology classes. However, since the selected path traverses through texture-poor region, it resulted in a higher odometry error and a lower success rate (3/10) compared to the proposed method. For sampling-based planners, since the environment is even larger and more complicated than the previous one, sample efficiency dropped and required more samples to find a visually advantageous path.  With 1000 and 8000 samples, the planner often selected path traversing texture-poor regions. Sampling-based planner with abundant sample number (15000) generated paths with lowest estimation error, but it took over 100 $s$ to generate the path at each trial. On the contrary, the proposed algorithm generated paths traversing texture-rich area in all trials, within a much shorter computation time. Results of each simulation environment are summarized in Table \ref{table_result}.

\begin{table}[tb]
  \caption{ \small{Simulation results for \textit{storage} and \textit{gallery} environments. Mean values of length of the groundtruth trajectory, distance between the goal and the estimated goal position, and computation time are measured. Note that goal estimation error was  evaluated only for successful trials.}} \label{table_result}
    \centering
    \begin{tabular}{|m{1.6cm}|m{1.0cm}|R{1.0cm}|R{1.3cm}|R{1.5cm}|}
    \hline 
 &Planner&Length~~&Goal Error&Comp. Time\\
\hline
\multirow{5}{*}{\hspace{0.35cm}\textit{storage}} & Proposed & $15.9$ $m$ & $0.118$ $m$ & $1.58$ $s$ \\
\cline{2-5}
& AP & $14.5$ $m$ & $0.391$ $m$ & $0.402$ $s$ \\
\cline{2-5}
 & R-300 & $9.86$ $m$ & $0.083$ $m$ & $3.04$ $s$ \\
\cline{2-5}
($12m \times 10m$)& R-1000 & $13.0$ $m$ & $0.136$ $m$ & $9.31$ $s$ \\
\cline{2-5}
 & R-2000 & $14.3$ $m$ & $0.082$ $m$ & $19.5$ $s$ \\
\hline
\multirow{5}{*}{\hspace{0.35cm}\textit{gallery}} & Proposed & $35.8$ $m$ & $0.235$ $m$ & $7.92$ $s$ \\
\cline{2-5}
& AP & $33.9$ $m$ & $0.664$ $m$ & $1.66$ $s$ \\
\cline{2-5}
& R-1000 & $36.8$ $m$ & $0.619$ $m$ & $9.52$ $s$ \\
\cline{2-5}
($22m \times 20m$)& R-8000 & $36.4$ $m$ & $0.598$ $m$ & $90.2$ $s$ \\
\cline{2-5}
& R-15000 & $36.7$ $m$ & $0.163$ $m$ & $164.1$ $s$ \\
\hline
    \end{tabular}
    \label{tab:my_label}
\end{table}

\subsection{Real-world Experiment}
%  To demonstrate the effectiveness of the proposed algorithm, real-world experiments were conducted. \textcolor{blue}{With Intel RealSense D435 camera, Pixhawk 4 flight controller, and Intel i7 NUC computer, a S-500 frame quadrotor MAV was set to move through consecutive three waypoints.} 
 For experiments, S-500 frame quadrotor equipped with forward-facing Realsense D435 camera was used. We used Pixhawk4 flight controller and Intel i7 NUC computer. 
 
\begin{figure}[tb]
      \centering
      \subfloat[\label{figure_experiment_env}]{
  \includegraphics[width=0.9\linewidth]{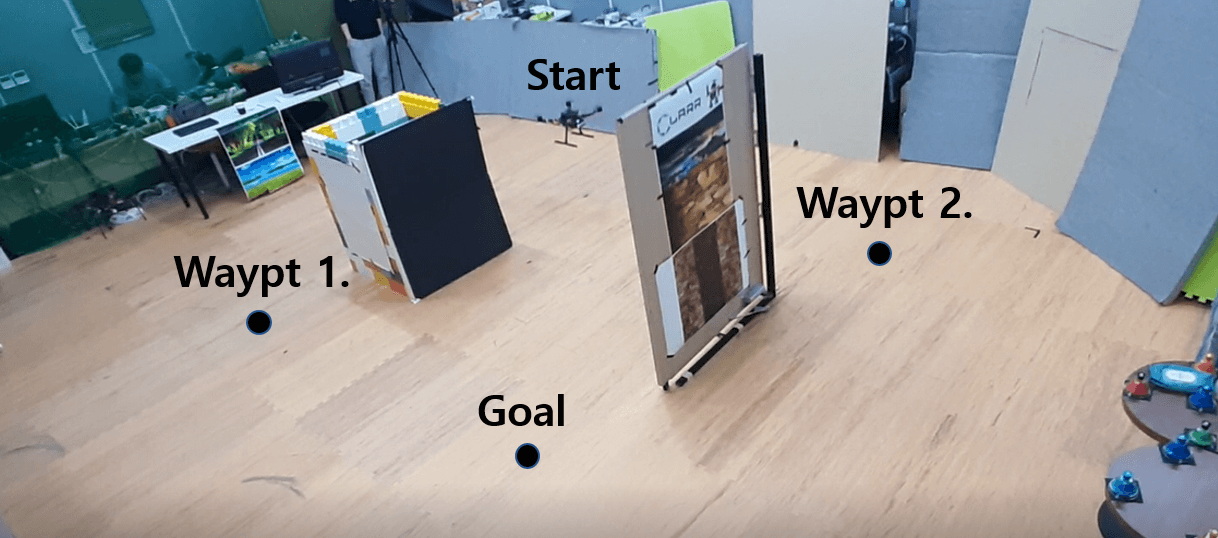}}
  \\ \vspace{-0.1cm}
    \subfloat[\label{figure_experiment_result}]{
  \includegraphics[width=0.95\linewidth]{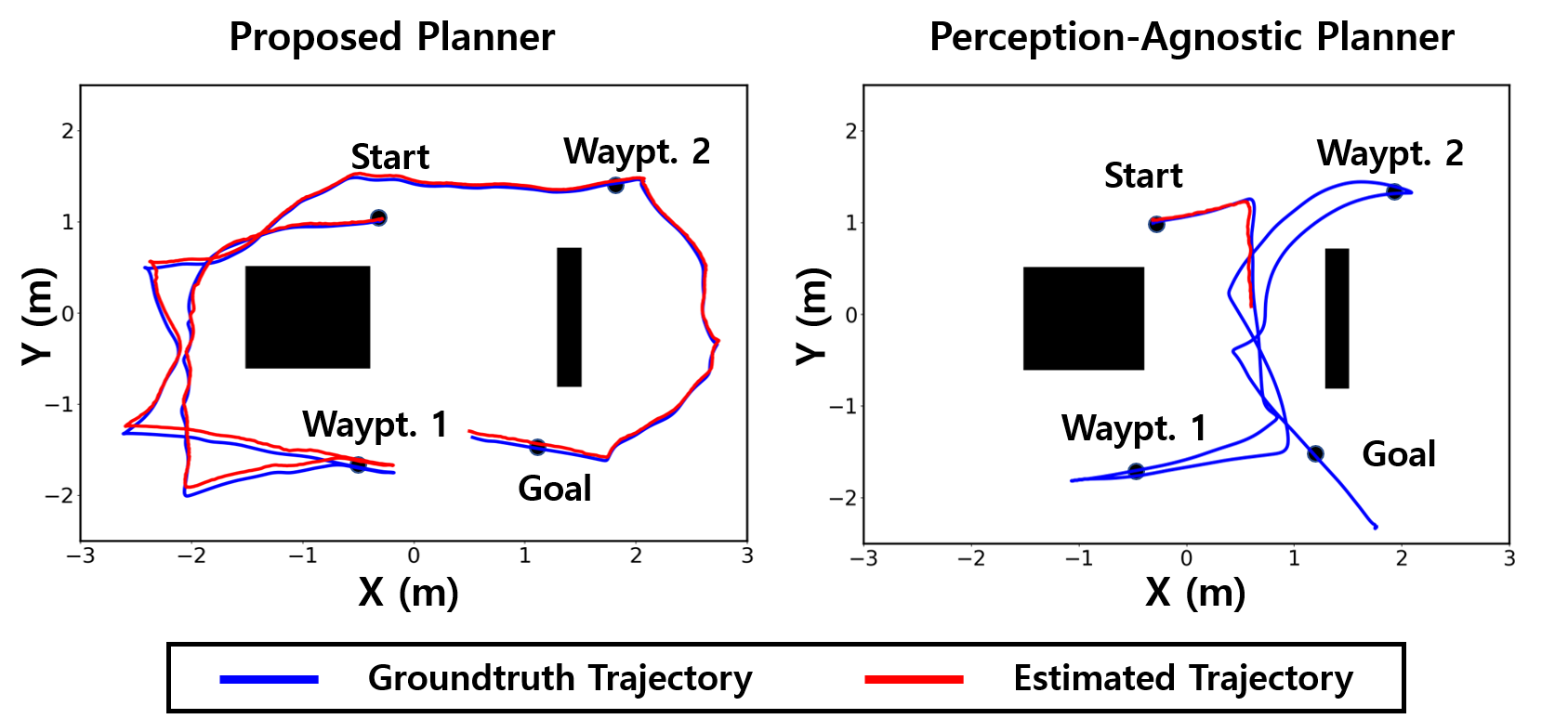} }
    \caption{
   Result from hardware experiments. (a) MAV generates a path pass to move through two waypoints and arrives at the goal point. (b) The resulting trajectories with the proposed planner (left) vs. the perception-agnostic planner (right). Blue lines indicate the groundtruth trajectory of MAV, and red lines are estimated position trajectory. With the perception-agnostic planner, the position estimate could be obtained during a part of the trajectory only, because of the early VO failure. }
   \label{figure_experiment}
   \end{figure}

As in Fig. \ref{figure_experiment}(a), the environment has two obstacles parallel to each other, with only a small number of features visible in the middle region and feature-rich objects visible from the rear sides. Two algorithms, the proposed algorithm and the perception-agnostic planner were mounted on MAV. We used OptiTrack motion capture system to provide state information to the controller in order to prevent control failure. It also provided the groundtruth poses. Estimated trajectories from visual odometry were compared with the groundtruth trajectories. As Fig. \ref{figure_experiment}(b). shows, the proposed planner created a path with a low odometry error through the region, but the perception-agnostic planner showed failure on VO since it did not take the visual information of the environment into account. 

\section{CONCLUSIONS \& FUTURE WORKS}
\colorcomment{}
Based on the observation that the homology class affects the visual information which MAVs can obtain, we proposed a topological path planner for perception-aware navigation. By creating a sparse topological graph from 2D GVD, multiple paths within distinctive homology classes are obtained. Then, the best path within each homology class is searched using graph search. Finally, among the paths selected within each homology, the path with minimum travel cost and perception cost is selected. We validated the effectiveness of our planner in multiple simulation environments and experiment. Future works would include an extension to online planning in unknown environments, reducing the need of tuning parameters (i.e.  perception quality threshold $c_{p, thr}$) via data-driven methods.

%%%%%%%%%%%%%%%%%%%%%%%%%%%%%%%%%%%%%%%%%%%%%%%%%%%%%%%%%%%%%%%%%%%%%%%%%%%%%%%%

% \section*{ACKNOWLEDGMENT}

%%%%%%%%%%%%%%%%%%%%%%%%%%%%%%%%%%%%%%%%%%%%%%%%%%%%%%%%%%%%%%%%%%%%%%%%%%%%%%%%

% \bibliographystyle{./bibtex/IEEEtran}
% % \bibliography{./bibtex/IEEEabrv, ./bibtex/sample} % from https://www.overleaf.com/project/5e453003ef93270001aaab70
% % \bibliography{./bibtex/IEEEabrv, ./bibtex/IEEEexample} % from http://ras.papercept.net/conferences/support/tex.php
% \bibliography{./bibtex/IEEEabrv, ./bibtex/mybibfile} % from my IROS 2019 paper


%%%%%%%%%%%%%%%%%%%%%%%%%%%%%%%%%%%%%%%%%%%%%%%%%%%%%%%%%%%%%%%%%%%%%%%%%%%%%%%%

\end{document}